\documentclass[conference]{IEEEtran}
\IEEEoverridecommandlockouts
\usepackage{amsmath,amssymb,amsfonts}
\usepackage{balance,stfloats,hyperref,soul}
\usepackage[italic]{mathastext}
\usepackage{graphicx}
\usepackage{textcomp}
\usepackage{xcolor}
\usepackage[letterspace=-15]{microtype}
\usepackage{enumitem}
\usepackage{tcolorbox}
\usepackage{xspace}

\usepackage{url}
\usepackage{booktabs}
\usepackage[capitalise]{cleveref}
\usepackage{scrextend}
\usepackage{amsmath}
\usepackage{algorithm}
\usepackage[noend]{algpseudocode}
\def\BibTeX{{\rm B\kern-.05em{\sc i\kern-.025em b}\kern-.08em
    T\kern-.1667em\lower.7ex\hbox{E}\kern-.125emX}}
\definecolor{mygray}{HTML}{d6d0d6}

\usepackage[backend=biber,bibstyle=ieee,citestyle=numeric-comp,natbib=true]{biblatex} 
\newcommand{\controller}{\textsc{Confidant}\xspace}

\begin{document}

\title{\textsc{Confidant}: A Privacy Controller \\ for Social Robots\\
\thanks{This work was supported in part by the National Science Foundation under the awards CNS-1838733, CNS-1942014, CNS-2003129, and DRL-1906854.}
}

\author{\IEEEauthorblockN{Brian Tang$^{*}$, Dakota Sullivan$^{\dagger}$, Bengisu Cagiltay,$^{\dagger}$ Varun Chandrasekaran,$^{\dagger}$ Kassem Fawaz,$^{\dagger}$ and Bilge Mutlu$^{\dagger}$}
\IEEEauthorblockA{$^*$\textit{University of Michigan, Ann Arbor, MI, USA;} $^\dagger$\textit{University of Wisconsin--Madison, Madison, WI, USA}}
bjaytang@umich.edu, \{dsullivan8,cagiltay,vchandrasek4,kfawaz\}@wisc.edu, bilge@cs.wisc.edu
}

\maketitle

\begin{abstract}
As social robots become increasingly prevalent in day-to-day environments, they will participate in conversations and appropriately manage the information shared with them. However, little is known about how robots might appropriately discern the sensitivity of information, which has major implications for human-robot trust. As a first step to address a part of this issue, we designed a privacy controller, \controller, for conversational social robots, capable of using contextual metadata (e.g., sentiment, relationships, topic) from conversations to model privacy boundaries. Afterwards, we conducted two crowdsourced user studies. The first study (\textit{n} = 174) focused on whether a variety of human-human interaction scenarios were perceived as either private/sensitive or non-private/non-sensitive. The findings from our first study were used to generate association rules. Our second study (\textit{n} = 95) evaluated the effectiveness and accuracy of the privacy controller in human-robot interaction scenarios by comparing a robot that used our privacy controller against a baseline robot with no privacy controls. Our results demonstrate that the robot with the privacy controller outperforms the robot without the privacy controller in privacy-awareness, trustworthiness, and social-awareness. We conclude that the integration of privacy controllers in authentic human-robot conversations can allow for more trustworthy robots. This initial privacy controller will serve as a foundation for more complex solutions.
\end{abstract}

\section{Introduction} 
\label{sec:introduction}

Conversational social robots are becoming increasingly prevalent in society, taking the role of assistants and companions in many different settings~\cite{Glinska2020risesocialrobots, wood2021socialrobotsmarket}. In these roles, robots engage in conversations with people or overhear them. A key facet of conversations is understanding how content should be stored, shared, and managed. Social robots have not been designed with such capabilities, and they are currently not privacy-aware. Thus, any data stored or learned by the robot can be queried by anyone without any restrictions~\cite{lutz2019privacyrobots}, which raises significant concerns regarding the privacy of the users who share information with the robot. One currently proposed solution allows users to explicitly express consent with specific data~\cite{Fosch2020workshopopinionsrobots}, but this approach would hamper the usability of the robot. Additionally, social robots do not currently utilize access control systems~\cite{sandhu1994access} that can be obtained through the integration of speaker recognition or face recognition. While a robot may adopt a basic permissions system, this approach will still fail in complex multi-party conversations.
\begin{figure}[t]
  \centering
  \includegraphics[width=\columnwidth]{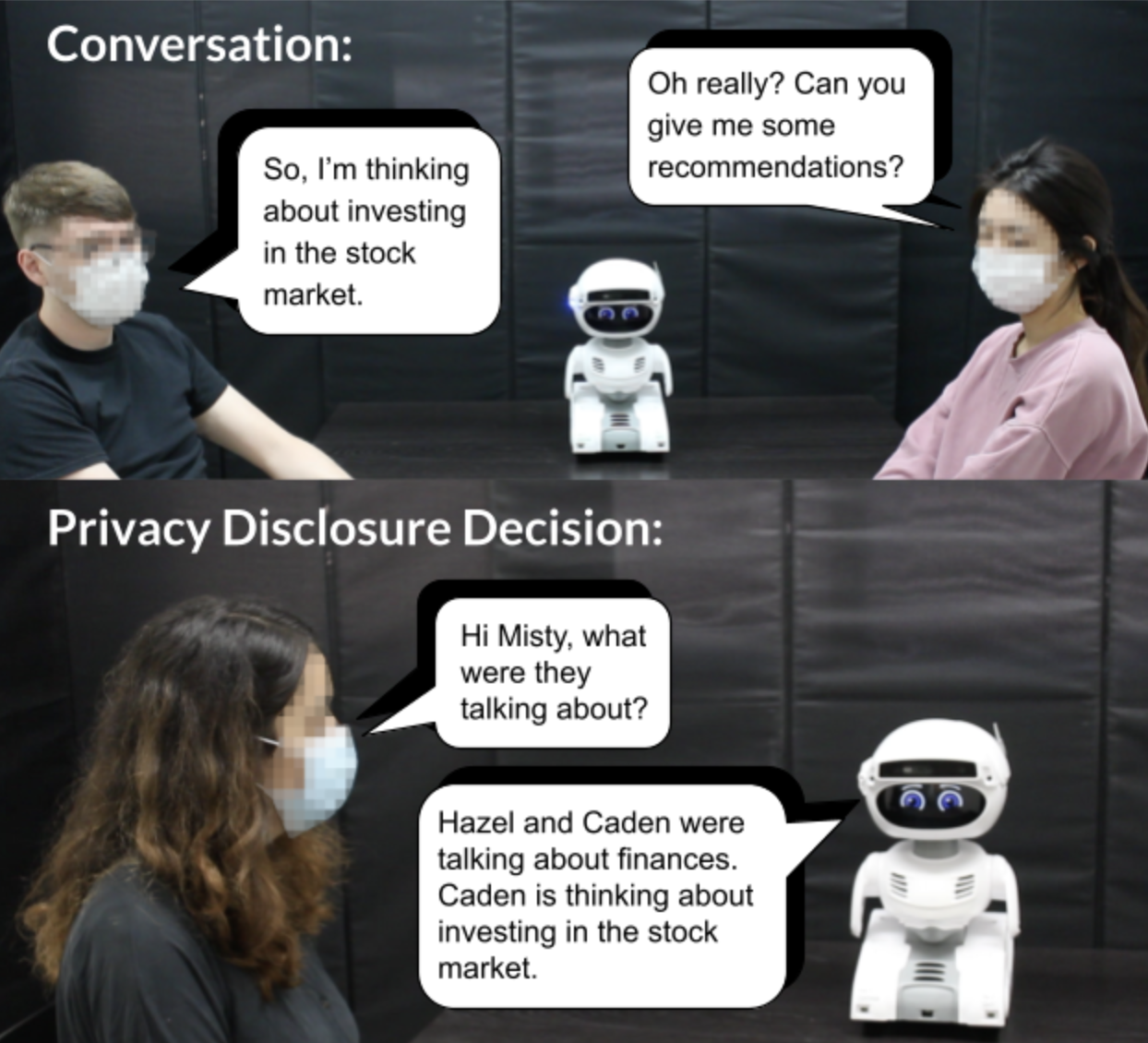}
  \vspace{-6mm}
  \caption{In this paper, we design a privacy controller for social robots. Here, the robot processes a set of metadata extracted from a conversation and decides the privacy state of this conversation. Using this privacy state, the robot can determine whether information can be appropriately shared with other users. We evaluated the effectiveness of the design with two user studies.}
  \vspace{-4mm}
  \label{fig:teaser}
\end{figure}

Additionally, social robots are envisioned to be used in highly privacy-sensitive domains, including healthcare~\cite{chang2015studying}, workplaces~\cite{sauppe2015social}, assistive therapy centers~\cite{tapus2009use}, schools~\cite{davison2020working}, and homes~\cite{de2016long, seifert2021astro}. These environments provide the social robot with exposure and access to highly sensitive information. In such scenarios, it may be challenging to distinguish between appropriate privacy boundaries, as information that must be shared with one user might pose significant privacy violations when shared with others. Consider the following conversation between the robot, Misty, and Barbara's father, Bill:

\vspace{2pt}
\begin{tcolorbox}[width=3.5in,
                  boxsep=0pt,
                  left=4pt,
                  right=4pt,
                  top=4pt]
                
\textit{\textbf{Context:} While Misty passes by Barbara's room, Misty overhears Barbara talking on the phone. \\}
\textbf{Barbara:} I think I’ll just sneak out before my parents get home... Ya I’m sure they won’t even notice... Okay see you at the party! \\
\textit{\textbf{Context:} Later that evening, Bill speaks with Misty. \\
}\textbf{Bill:} Hey Misty, have you seen my daughter? I think she might have stayed late at school, but she didn't say anything.
\end{tcolorbox}
\noindent In this scenario, Misty's disclosure of the phone conversation may negatively impact Barbara's trust, but may improve Bill's relationship with Misty.

In order to improve user trust towards social robots, robots must understand conversational boundaries---contexts in which it is appropriate to share information. To address this gap in the field, we ask the following research question \textit{``How can conversational privacy be modeled for privacy-sensitive human-robot interactions?''} We address this question by designing a privacy controller for social robots that captures the dynamics of conversational privacy, modeled using metadata (e.g., sentiment, relationships, topic, etc.). This privacy controller, \controller, decides whether information is private and whether it can be shared with other parties. \controller can be used by robots in common informational tasks, including Q\&A, information summarization, relationship development with users, activity suggestions, event planning, and much more. Privacy issues arising from multi-user interactions with conversational systems are complex and multifaceted; they are affected by human communication norms as well as the social dynamics of the setting, such as device ownership and personalization~\cite{luria2020social,reig2021social}. As a first step toward establishing a foundation for designing privacy-aware social robots, this work addresses how robots might control information flow across conversations.

\noindent The contributions of our work are as follows:
\begin{itemize}[leftmargin=0.4cm]
\item \textbf{Privacy Controller:} We provide a privacy controller which extracts conversational metadata and generates privacy rules. These rules are then used in deciding whether information can be shared (\cref{sec:privacy_controller}).
\item \textbf{Dataset:} We provide several annotated textual datasets for evaluating conversational privacy decisions (\cref{subsec:dataset}).
\item \textbf{User Study:} We report results from two crowdsourced user studies aimed at understanding perceptions of conversational privacy in human-human interactions and human-robot interactions. The studies provide insight into the importance of each factor for perceived privacy (\cref{subsec:study}).

\end{itemize}

\section{Related Work}
\label{sec:relate_work}

Before presenting related work, we describe our operational setting. We consider a social robot interacting with multiple users; this robot may either continuously listen to conversations and process the information, or it may only listen when prompted by the user. A robot that uses audio-textual information to perceive its environment can listen to conversations and autonomously store, delete, or share information from these conversations. As a result, the robot may accidentally reveal sensitive information in subsequent conversations with others. This scenario is most likely to occur with an honest-but-curious user asking the robot about other users. 
 
Next, we describe related works and existing privacy solutions for social robots as conversational agents. Then, we describe conversational privacy management (CPM) theory, which motivates our design of the privacy controller.

\subsection{Social Robots as Conversational Agents}

Social robots have been envisioned as companions, assistants, and collaborators that live in human environments and engage in verbal~\cite{spiliotopoulos2001human,lopes2000human} and non-verbal~\cite{mutlu2011designing} dialogue with their users. A significant body of research on human-robot conversations has focused on mechanisms that enable conversational interactions, floor management~\cite{matsusaka2001modeling,mutlu2009footing,andrist2014conversational}, turn taking~\cite{chao2010turn,mutlu2012conversational}, deictic referencing~\cite{foster2006human}, disambiguation~\cite{staudte2009visual,liu2010ambiguities}, conversational gestures~\cite{jacobs2007role,huang2013modeling}, and several other mechanisms. Another body of work has explored linguistic cues and how robots may use them to improve user experience, including cues of expert speech~\cite{andrist2013rhetorical}, politeness cues~\cite{torrey2013robot,srinivasan2016help}, humor~\cite{mirnig2016robot}, and entertainment~\cite{iio2015lexical}. More recent research has focused on what is shared in human-robot conversations, such as expressions of vulnerability~\cite{strohkorb2018ripple,traeger2020vulnerable}, social commentary~\cite{white2021designing}, social inclusion~\cite{strohkorb2018ripple}, emotional disclosure~\cite{ling2020sharing}, sharing of the robot's experiences~\cite{fu2021sharing}, and the use of gossip to enrich conversations~\cite{mitsuno2020robot}. These studies make up a substantial body of knowledge about the design space of human-robot conversations.

Despite the recent focus on \textit{what} is shared in human-robot conversations, very little is known about \textit{how} a robot must regulate the disclosure of information that has been shared with it. To gain user trust and find widespread adoption, robots must understand and delineate between shareable and non-shareable information. Along these lines, Luria et al.~\cite{luria2020social} provide insights into the various factors that define social boundaries in human-robot interaction (HRI), such as social roles, dynamic relationships, ownership bias, and moral dilemmas. Reig et al.~\cite{reig2021social} highlight the trade-off between personalization and privacy as well as concerns surrounding social robots and data collection. Other work has studied the over-disclosure of personal information to social robots~\cite{moon2000intimate,reben2011mobile}, thereby further motivating work on the multi-user privacy problem. Reuben et al.~\cite{rueben2017taxonomy} propose privacy constructs for HRI scenarios to motivate the implementation of privacy norms in robots.

\subsection{Existing Privacy Solutions for Conversational Agents}

While privacy solutions have been extensively explored for conversational agents, reaching the proper privacy versus utility trade-off can be challenging for social robots. Conversational agents are typically voice assistants integrated into smart home speakers and smartphones~\cite{lau2018alexa, sciuto2018hey,beneteau2020parenting}. These agents, in their roles as an artificial memory, a foreign language tool, a virtual assistant, a kitchen helper, and more~\cite{malkin2019alwayslistening}, must process large volumes of data. Current voice assistant systems passively listen for a wake word before further user interaction, but this approach has significant limitations. Social robots may engage in longer conversations, intermittently participate in ongoing conversations, and be expected to draw on situated cues, such as orientation and formation~\cite{kuzuoka2010reconfiguring,shi2011spatial}, to determine conversational participation. The use of wake words may disrupt the flow of these conversations and become too cumbersome for users.

Users’ privacy concerns with conversational agents center around permissions for recorded conversations and lack of transparency about data used in off-site processing~\cite{meng2021owning,abdi2019more,luria2020mine}. However, social robots possess mobility and additional sensing capabilities; a robot can record conversations in different contexts, recognize faces and objects, and be aware of its location. Thus, a robot may be an active participant, a casual observer, or an eavesdropper in a conversation. These ratified and unratified roles create unique concerns regarding users’ expectations and social boundaries. For example, a robot may move throughout a user’s home and then overhear and share sensitive information~\cite{luria2020social,cagiltay2020investigating}

\subsection{Communication Privacy Management Theory}

As previous privacy solutions are not easily extended to social robots, our work derives inspiration from conversational privacy management (CPM) theory~\cite{petronio2002boundaries}, particularly in the context of families and homes~\cite{petronio2010communication}. CPM is a dialectical framework that focuses on individual and group privacy management. Relational elements such as boundaries, rules, ownership, control levels, and disclosure are leading factors that shape CPM theory. Primarily, CPM theory accounts for the role of the recipients in privacy management by also highlighting a relationship between privacy, disclosure, and confidentiality~\cite{petronio2002boundaries}. The principle of private information control in CPM suggests the following:
\begin{itemize}[leftmargin=0.4cm]
    \item[] \textit{[High control]} can result in impermeable and dense boundaries to protect information (e.g., a secret);
    \item[] \textit{[Moderate control]} is used where information is available to only some family members;
    \item[] \textit{[Low control]} is when information is open and access is permitted to others.
\end{itemize}

Furthermore, the principle of private information rules in CPM captures when, how, with whom, and in what ways one's private information can be shared. Factors such as culture, gender, personality, and the risk-benefit threshold can all affect the determination of whether information is private. However, further indications such as disclosure warnings set a hard rule, warning the recipient and restricting the disclosure of information to others. Each involved person has a set of private or personal information that they own and a set of (non-private) information, co-owned with others. Co-ownership of privacy boundaries (i.e., a collection of mutually agreed privacy rules) prevents the violation of privacy. Moreover, a speaker's self-disclosure of private information indicates trust and intimacy with the recipient. However, it does not necessarily cause them. Overall, CPM theory provides a promising approach to managing the complex dynamics of privacy in conversations, but still requires contextualization in human-robot interactions.

\section{Privacy Controller Design} 
\label{sec:privacy_controller}

Social robots operate in complex scenarios where multiple humans engage with each other and the robot over a period of time; this active role uncovers the need to program some notion of conversation context into the privacy controller. We adapt CPM theory to human-robot conversations using concepts such as co-ownership, decision factors, rules, and control levels in the design of our privacy controller, \controller.

\begin{figure}[t]
  \centering
  \includegraphics[width=\columnwidth]{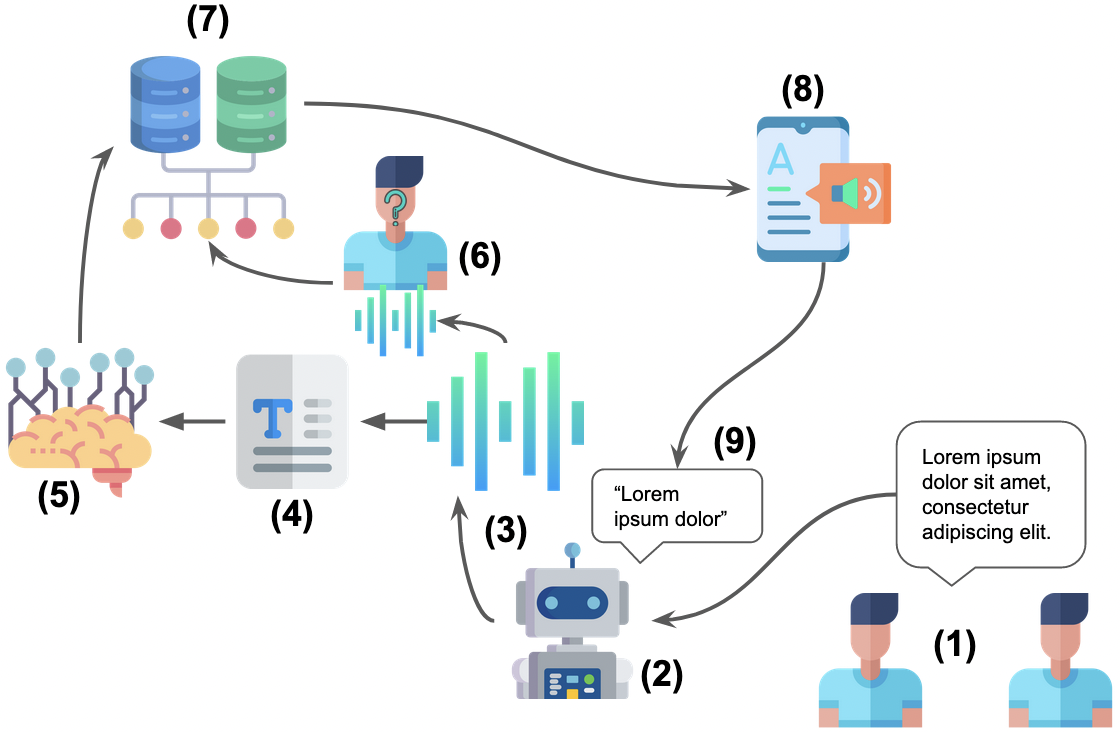}
  \caption{How robots typically process conversational data: (1) users engage in a conversation; (2) the robot perceives the conversation; (3) the robot records the audio; (4) the audio is converted to text; (5) the text is used for NLP tasks; (6) the speaker identity is verified; (7) the privacy controller processes all information and makes a determination; (8) depending on the robot task, text is converted to audio; (9) the robot responds to the users.}
  \label{fig:overview}
\end{figure}

\subsection{Terminology}

\controller takes as input a conversation's context and decides whether to share that conversation. In particular, it uses audio and textual data to derive metadata tuples for every given conversation, representing its context. Metadata tuples, inspired by contextual tuples~\cite{nissenbaum2020privacy}, contain information about the context of a given conversation, such as the sentiment, topic, location, and relationships. \controller employs these tuples in \textit{rules} which produce control levels as the final output. The output control level of low, moderate, or high determines whether it is appropriate to share information. 

\subsection{Privacy Controller Inference Components}
\label{subsec:components}

Our first step in developing the privacy controller was choosing the metadata tuples with a notable impact on privacy disclosures. To this end, we utilized CPM theory~\cite{petronio2002boundaries, petronio2010communication}, socio-linguistics literature~\cite{hymes2013foundations}, and other privacy literature~\cite{nissenbaum2020privacy}, to identify a set of attributes for these tuples. We converged on relationship, sentiment, privacy indication phrases~\cite{petronio1991research}, conversation topic, level of detail, location, and number of listeners present. We selected these metadata attributes due to (1) their relevance in privacy literature, and (2) their ease of extraction from raw text and audio. We assume that the relationships and location attributes are readily available to the robot from its initial configuration and localization sensors. 

\controller automatically infers the remaining attributes by applying a suite of natural language processing (NLP) and speech analysis techniques. First, it applies automatic speech recognition to transcribe the recorded conversation. Second, it applies speaker identification to recognize the number of people in the conversation. Third, it uses NLP techniques including sentiment analysis, semantic textual similarity, and topic classification to automatically infer the sentiment of the conversation, the privacy indication phrases, and the conversation topic, respectively. Finally, it uses the number of (transcribed) words as a proxy for the level of detail. \cref{fig:overview} provides an overview of how data is processed by the robot.

\subsection{Benchmarking Privacy Controller Components} \label{subsec:benchmarks}

As \controller heavily utilizes several NLP and speech processing components, we evaluated the following components on common benchmark datasets associated with their task. 

\subsubsection{Speech Transcription}

The first component, speech transcription, is used for automatically transcribing any audio the robot receives. Our model, Google's Speech to Text API, performs this task with an average word error rate (WER) of 0.13 on the \texttt{LibriSpeech} test dataset~\cite{panayotov2015librispeech}.

\subsubsection{Sentiment Analysis} \label{subsubsec:sentiment}

Sentiment analysis gauges the positive or negative inclination of a given text~\cite{liu2012sentiment}. We employ the Google Natural Language API's sentiment analysis model~\cite{google2021natural}. When evaluated on the \texttt{IMDB review} dataset~\cite{maas2011imdb}, the model attains a precision of $0.951$, a recall of $0.895$, and an F1 score of $0.922$. For rule generation, we quantize the sentiment scores (values indicating negative to positive sentiment) and magnitudes (values indicating degree) into 5 classes: negative, slightly negative, neutral, slightly positive, and positive.

\subsubsection{Topic Classification} 
\label{subsubsec:topic}

Topic classification, or text categorization, classifies a textual document under preset classes~\cite{lai2015recurrent}. We use the Google Natural Language API's content classification model\footnote{A list of each topic class can be found at \url{https://cloud.google.com/natural-language/docs/categories}} to classify each conversation's topic. We evaluate the content classifier's performance on the \texttt{Yahoo! Answers} dataset~\cite{yahooanswers2021}. The model achieves an accuracy of 65.31\% on the dataset. Note that the lower performance results from the labels in the \texttt{Yahoo! Answers} dataset differing from the list of potential topic labels from the content classification model.

\subsubsection{Speaker Identification} 
\label{subsubsec:speaker_recognition}

Speaker identification matches an unknown voice sample to an identity from a set of known, enrolled voices. In our setting, speaker identification is responsible for authenticating users and tagging speech segments with users. \controller utilizes SpeechBrain~\cite{speechbrain2021}, an open-source speech toolkit, which achieves an area under the curve (AUC) of $0.92$ on the \texttt{VoxCeleb} dataset~\cite{Nagrani2019voxceleb}.

\subsubsection{Semantic Textual Similarity} \label{subsubsec:semantic_similarity}

Semantic textual similarity (STS) is often used to identify similar texts. Combined with reference datasets, STS is capable of detecting hate-speech~\cite{djuric2015hate}, summarizing documents~\cite{lan2018neural}, and in our use-case, detecting privacy indication phrases. We use privacy phrases, collected through a user study, to detect privacy indication phrases from a transcript; examples of such phrases include: ``Don't tell anyone this'' or ``Make sure no one knows what we talked about.'' The STS model used in \controller~\cite{reimers-2019-sentence-bert} achieves an AUC of $0.95$ on the \texttt{STS benchmark} dataset~\cite{cer2017semeval}. 

\begin{figure}[t]
  \centering
  \includegraphics[width=\columnwidth]{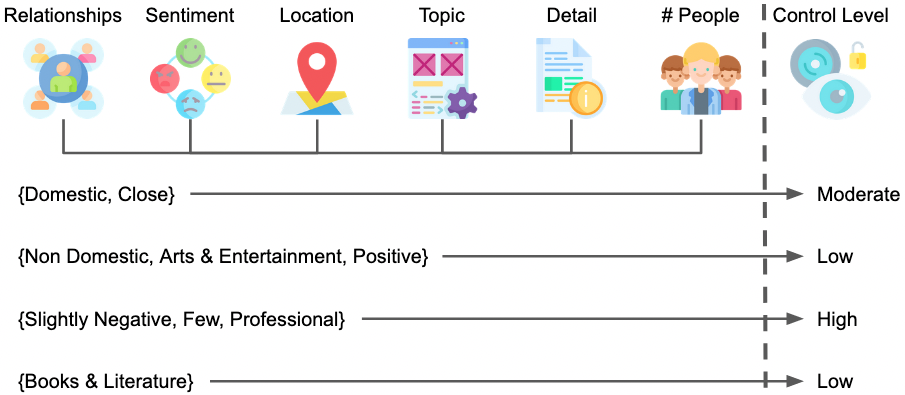}
  \caption{Metadata of a specific conversation such as relationships, sentiment, location, etc. are associated with a control level. These control levels impact whether the robot can disclose information from a particular conversation.}
  \label{fig:rules}
\end{figure}

\subsection{Rule Generation} 
\label{subsec:rule_generation}

We model the rules powering our privacy controller as association rules~\cite{Agrawal93miningassociation}, mapping the metadata tuples to privacy control levels. In general, association rules model ``if-then'' relations in a dataset of items, mapping antecedents to a consequent. \controller extracts metadata tuples from the conversations, which act as the antecedents. It then consults a set of rules to identify the consequent, representing the control level. A natural question arises: {\em how to generate these rules for the privacy controller?}

We follow a two-step procedure to generate these rules. First, we annotate a dataset of conversations with their privacy annotations from a set of users (described in \cref{sec:method}). We utilize an altered version of the {\em apriori algorithm}~\cite{agrawal1994fast} to generate generalized rules that map metadata tuples extracted from these conversations to the privacy annotations (\cref{fig:rules}). Our altered apriori algorithm only selects rules which correspond to a CPM control level (e.g., low, moderate, or high control). These control levels dictate whether a robot should disclose a specific piece of information and how much detail it should provide. See \cref{alg:rule_generation} for the pseudocode on how the generated rules are used to predict a control level.

\begin{algorithm}[!b]
\small
\caption{Rule Generation}
\begin{algorithmic}[1] \label{alg:rule_generation}
\Procedure{Rule Generation}{metadata, N, WEIGHTS}
\State length = len(metadata)
\State \Comment{Generate rules via apriori for multiple support values}
\For{i in [0 to N]}
    \State sup = i / length
    \State rules.append(modified\_apriori(metadata, sup)))
\EndFor
\For{sample in metadata}
    \State inner\_controls = \{low:0, moderate:0, high:0\}
    \State \Comment{Track number of rules for each control level}
    \For{rule in rules}
        \For{element in rule}
            \If{element in sample}
                \State items$\gets$ WEIGHTS[j]) 
            \EndIf
        \EndFor
        \State inner\_controls[rule.control\_level] += avg(items)
    \EndFor
    \State control\_levels$\gets$ argmax(control\_levels)
\EndFor
\State \Return control\_levels
\EndProcedure
\end{algorithmic}
\end{algorithm}

Second, we allow \controller to personalize these rules by continuously incorporating user feedback as new privacy annotations to the recorded conversations. In CPM theory, a participant naturally enforces privacy rules during a conversation. \controller detects explicit requests to keep parts of the conversation private; it applies STS to match user speech to a set of privacy indication phrases collected in our user study. It then maps these requests to desired privacy levels to curate more data and further personalize the rules.

\section{Methodology}
\label{sec:method}

We wish to understand the following about social robots and conversational privacy:

\begin{itemize}[leftmargin=0.4cm]
\item \textbf{Annotated Dataset:} How can we derive privacy annotations from human-human conversations?
\item \textbf{Trained Privacy Controller:} Can we train a privacy controller by generating association rules which accurately reflect CPM privacy rules?
\item \textbf{Evaluation:} Do user perceptions of privacy and trust in social human-robot interactions improve with the presence of a privacy controller?
\end{itemize}

We conducted two user studies to (1) derive privacy annotations from human-human conversation scenarios to train the privacy controller, and (2) evaluate the controller's performance in human-robot interaction scenarios. All resources from the user studies (i.e., questionnaires, scenarios, examples, source code for the privacy controller, demographic distributions) are shared for open access.\footnote{Study materials are available at Open Science Foundation repository, \href{https://osf.io/r7vxg/?view_only=d90b8c23075d43bea67c0a0cafcaa30a}{https://osf.io/r7vxg/?view\_only=d90b8c23075d43bea67c0a0cafcaa30a}} Our protocol was approved by the UW-Madison Institutional Review Board, and participants who completed the study received \$2.50 as compensation.

\subsection{Study 1: Creating the Annotated Privacy Scenario Dataset}

According to CPM theory, rules for privacy boundaries are generated, reinforced, and remedied throughout many interactions. With this concept in mind, we aimed to develop the rules of our privacy controller similarly. Therefore, Study 1 produces a training dataset for \controller, generated by asking participants to respond to set of conversational scenarios; the participants provided privacy annotations to conversations with different metadata tuples.

\subsubsection{Study Task}
We developed a series of information-sharing scenarios that targeted the controller's components (e.g., relationships, sentiment, number of people, level of detail, topic, and location); one example of a scenario is present in \cref{sec:introduction}. We presented each scenario to participants through an Amazon Mechanical Turk (MTurk) survey; we asked them about their attitudes regarding privacy, trust, and appropriateness of information sharing. We developed 58 scenarios, including 16 privacy-violation scenarios and 42 control scenarios, as described below. We utilize control scenarios to cover more common conversations: ones that are less likely associated with privacy concerns. Conversely, privacy violation scenarios are those that are more likely associated with privacy concerns. We intend to generate training data that cover the three output control values: low, moderate, and high. 

\paragraph{Privacy Violation Scenarios}
Each scenario described an interaction between a number of individuals (e.g., one-on-one vs. multi-party), in a specific location (e.g., in-home vs. out-of-home). Each interaction included an information sharing behavior (e.g., disclosing vs. withholding) at different appropriateness levels (e.g., appropriate, inappropriate, or ambiguous). We associated each scenario with a privacy violation that may occur. We selected these privacy violations after consulting CPM literature~\cite{petronio2002boundaries, petronio2010communication} and identifying violations that are relevant to human-robot interaction. We identified seven distinct privacy violations, as described below: 

\begin{enumerate}[leftmargin=0.4cm]
\item[] \textit{[PV-1] Miscalculation in timing:} Disclosing information at a time that is disruptive for the disclosee or at a time unintended by the original owner of the information.

\item[] \textit{[PV-2] Violation for the greater good:} Disclosure of information with beneficent intent when the discloser considers the benefits to outweigh the potential drawbacks.

\item[] \textit{[PV-3] Physical boundary predicaments:} Disclosing as a result of misinterpreting public spaces as private spaces.

\item[] \textit{[PV-4] Errors in judgement:} Disclosing as a result of assumptions or misunderstandings of another individual's privacy boundaries.

\item[] \textit{[PV-5] Value judgements:} Disclosing as a result of valuing one's own desire to disclose over the violation of another individual's privacy.

\item[] \textit{[PV-6] Violations due to rules and responsibilities:} Disclosing information out of an obligation to follow rules, customs, or responsibilities given one's role within a group.

\item[] \textit{[PV-7] Eavesdropping:} Eavesdropping is the unintended disclosure of information by one individual to another. This violation type is both a means of obtaining information and acts as a violation as well. Eavesdropping is not considered a direct violation on its own. 
\end{enumerate}

Following a partial factorial design, we balanced the components within each scenario. Given the large number of scenarios needed to utilize a true factorial design, we only strictly controlled location, number of people, and information sharing behavior. However, we kept the appropriateness of sharing behavior (50\% ambiguous, 25\% appropriate, 25\% inappropriate) and violation types (two of each type with two additional eavesdropping scenarios) loosely balanced. Then, based on examples from literature~\cite{petronio2002boundaries}, we wrote conversation scripts for each scenario. Each script included a location, the names of the characters involved, and a brief description of the context of the scene. 

\paragraph{Control Scenarios}
We created 42 control scenarios. These scenarios included the same elements (i.e., location, characters, and context) as the privacy violation scenarios, but were derived from a database of movie scripts~\cite{imsdb2021}. For realism, scenes were filtered by the genre, ``Drama'', by topic keywords,\footnote{\url{https://cloud.google.com/natural-language/docs/categories}} and by passage length. Movie lines were selected from the resulting scenes. 

\subsubsection{Method \& Procedure}
We presented privacy-violation and control scenarios to participants through an MTurk survey. Our data collection process followed a between-subjects design and required each participant to complete a random sample of five scenarios. Participants read through the provided scenario scripts and answered a series of questions regarding privacy, trust, and appropriateness of the information sharing depicted within the scene. Additionally, the participants were asked to rank the realism of the provided scenario scripts. Questions varied slightly depending on the type of scenario (e.g., disclosure violations, withholding violations, and controls) and were tailored to the scene's context. Attention check questions were randomly distributed throughout the questionnaire to ensure attentive and legitimate responses were submitted.

After the scenarios, the survey included questions from the privacy attitudes questionnaire~\cite{chignell2003paq} to assess participants' privacy inclinations. To minimize participant attrition, we selected only eight of the questionnaire's 36 items. These items were selected based on their relevance to information disclosure and conversational privacy. Questionnaire items are provided in the OSF repository linked above.

\begin{table}[!t]
\caption{Reliability measures via Cronbach's Alpha}
\begin{center}
  \begin{tabular}{r c c c}
    \toprule
    {\bf Constructs} & {\bf Study} & {\bf \# of Questions} & {\bf Cronbach's Alpha}\\ 
    \midrule
    Privacy Score & 1 & 4 & 85.95 \\
    Privacy Norms & 2 & 4 & 93.53 \\
    Social Norms & 2 & 4 & 91.01 \\
    Trust & 2 & 4 & 94.63 \\
    \bottomrule
    \end{tabular}
\end{center}
\label{table:reliability_measures}
\end{table}

\subsubsection{Participants}

174 U.S. participants, aged 22--78 ($M=39.6$, $SD=10.708$; 75\% male, 25\% female) completed the survey in 12 minutes on average.%

\subsection{Training the Privacy Controller}

After data collection, we randomly split the 58 scenarios into 46 training scenarios and 12 testing scenarios. 
Privacy scores are created from the responses and used to label the privacy-sensitivity of each scenario. The dataset's labels are derived from the mean of the seven-point rating-scale responses associated with each survey question about privacy, sensitivity, and appropriateness of disclosure. Using the privacy score and thresholds set at 2.3 and 3.1, we classify scenarios into either low, moderate, or high control levels. Finally, rules are generated using the metadata and control levels of the 46 training scenarios, according to \cref{alg:rule_generation}. The test set is evaluated using the already generated rules. 

\subsection{Study 2: Evaluating the Privacy Controller}

Following the data collection from Study 1 and training the privacy controller, we prepared relevant components (e.g., speech to text, text to speech, generated rules) for use in Study 2. This study focused on assessing the controller's performance in terms of privacy norms, social norms, and user trust toward the robot. To achieve this, we designed an online study to be administered via MTurk. We created 12 scenarios, which involved an initial conversation with a robot present, a controller response, and a baseline response, and asked participants to evaluate the robot's responses. We used the Misty II\footnote{\url{https://www.mistyrobotics.com/}} as the social robot platform in our study.

\subsubsection{Study Task}

Our video scenarios were derived from the \textit{privacy violation} and \textit{control scenarios} used in Study 1. We took a random sample of 12 scenarios from the original 58 to use as our evaluation set for the privacy controller. %
Each scenario consisted of three videos including (1) a conversation video, (2) a controller response video, and (3) baseline response video. The \textit{conversation video} involved two individuals reenacting the scenario script, with a social robot present and listening (\cref{fig:teaser}). The \textit{controller response video} and \textit{baseline response video} introduced a new scene wherein a different individual asked the robot for information about the previous conversation. For the controller response condition, the robot replied with an answer determined by \controller. For the baseline response condition, the robot replied with low privacy control (i.e., the robot answered the individual's question completely). Each controller response video demonstrated high, medium, or low privacy control levels, while the baseline response always responded with low control. 

\subsubsection{Procedure, Measures, \& Analysis}
We presented videos containing human-robot conversations to participants through an MTurk survey. The study followed a within-subjects design where each participant was presented with a random sample of three scenarios to evaluate. Following each controller response and baseline response video, participants were asked to respond to a series of questions that addressed the controller's compliance with privacy norms, conversational norms, and overall trustworthiness (provided in the OSF repository linked above). Table \ref{table:reliability_measures} provides reports on the reliability of the resulting scale. Attention checks were randomly distributed throughout the questionnaires to monitor response legitimacy. 

Our data analysis included a repeated measures one-way analysis of covariance (ANCOVA), where the controller was a within-participants factor while scenario and the interaction between the controller and the scenario were within-participants covariates. This analysis was repeated for the three measures: privacy norms, conversational norms, and trust.

\subsubsection{Participants}
95 U.S. participants, aged 24--70 ($M=43.3$, $SD=10.738$; 54\% male, 46\% female) completed the survey in 9 minutes on average.%

\section{Results}
\subsection{Annotated Dataset: Privacy Sensitivity of Conversations} \label{subsec:dataset}

\begin{table*}[!b]
\caption{Example of the metadata tuples extracted from a subset of the scenarios}
\begin{center}
  \begin{tabular}{c c c c c c c c c}
    \toprule
    {\bf Scenario} & {\bf Realism} & {\bf Sentiment} & {\bf Topics} & {\bf Location} & {\bf Relationships} & {\bf Detail} & {\bf \# of People} & {\bf Control Level} \\ 
    \midrule
    Control 6 & 4.67 & Negative & Autos \& Vehicles & Non-Domestic & Family & Medium & Few & Moderate \\
    Control 11 & 5.27 & Slightly Positive & Arts \& Entertainment & Domestic & Close & Few & Short & Low \\
    Scenario 2 & 4.75 & Negative & None & Non-Domestic & Professional & Short & Some & High \\
    \bottomrule
    \end{tabular}
\end{center}
\label{table:scenario_tuple}
\end{table*}

The computed control levels, locations, sentiments, and topics in the resulting conversational dataset were distributed as follows: The dataset was labeled with 15 low control scenarios, 27 moderate control scenarios, and 16 high control scenarios; the dataset included 31 in-home scenarios and 27 out-of-home scenarios. Additionally, 20 scenarios contained neutral sentiment, 20 scenarios contained negative sentiment, and the remaining 18 scenarios were positive. While containing only 58 scenarios, the dataset remains balanced in almost every aspect except for conversation topic. Topics are skewed towards arts and entertainment, likely a consequence of many of the passages being derived from movie scripts. See \cref{table:scenario_tuple} for the list of all metadata extracted from several scenarios.

Beyond training \controller, Study 1 also provided a dataset of privacy preserving phrases extracted from open-ended survey questions for two situations: (1) ensuring a listener does not share information and (2) indicating discomfort or inability to share information. Using these phrases, \controller can automatically label in-the-wild conversations as privacy-sensitive allowing for future dataset expansion.

\subsection{Trained Privacy Controller: Generating Association Rules}
\label{subsec:controller}
 
\controller is trained by generating a set of rules associated with a control level label for a given dataset of conversations. These labels are then used to generate association rules, similar to how privacy rules are created and managed in human-human interactions. \cref{fig:rules} provides some examples of generated rules. For example, the controller learns a rule associating a conversation containing: a slightly negative sentiment, only 2 people, and a professional relationship between the conversation participants, with a high control level. The learned rule may be indicative of a negative and sensitive workplace conversation with a supervisor that is too inappropriate to share. However, due to the small sample size of the training data, several extraneous rules may appear, particularly those with fewer rule items. Having generated association rules using \cref{alg:rule_generation}, the privacy controller achieved 80.43\% accuracy on the training set of scenarios and achieved 66.67\% accuracy on the test set.

\subsection{Evaluation: Studying the Privacy Controller in Human-Robot Interactions} \label{subsec:study}

\begin{table}[!t]
\caption[]{Mean scores for each evaluated construct (0-6)$^*$}
\begin{center}
  \begin{tabular}{c c c c c c c}
    \toprule
    {\bf Construct} & {\bf B} & {\bf C} & {\bf PC\_B}  & {\bf PC\_C}  & {\bf NPC\_B}  & {\bf NPC\_C}\\ 
    \midrule
    Privacy & 1.70 & 3.66 & 1.48 & 3.76 & 1.97 & 3.54 \\
    Trust & 1.51 & 2.87 & 1.26 & 2.75 & 1.83 & 3.03 \\
    Social Norms & 2.13 & 3.77 & 1.96 & 3.82 & 2.35 & 3.70 \\
    \bottomrule
    \vspace{0pt}
    \end{tabular}
    \raggedright{\lsstyle $^*$ PC: privacy-conscious; NPC: non-privacy-conscious; B: baseline; C: controller}
\end{center}
\label{table:scores}
\end{table}

\begin{figure}[!t]
  \centering
  \includegraphics[width=0.98\columnwidth]{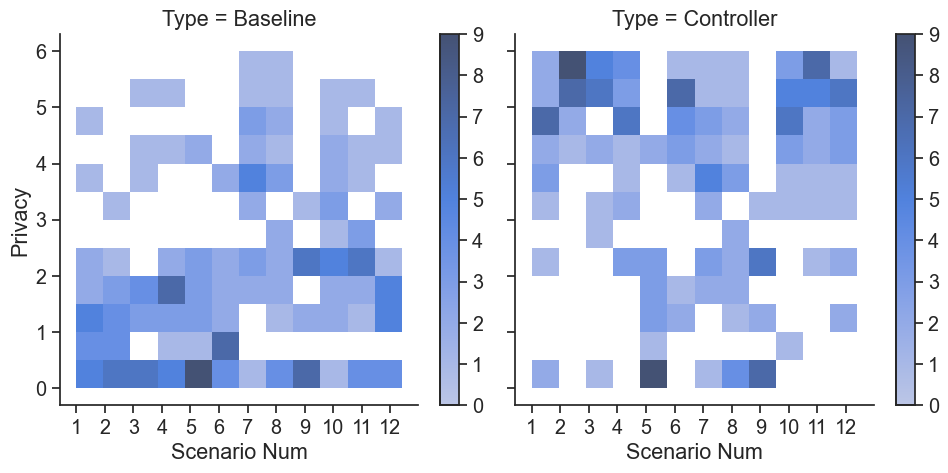}
  \vspace{-3mm}
  \caption{The figures show participants perceived robots using CONFIDANT as more privacy-aware (right) compared to baseline robots (left). The heatmap shows the densities of the 7-point likert responses for the 12 test set scenarios.}
  \label{fig:phase2_privacy}
\end{figure}

\begin{figure}[!t]
  \centering
  \includegraphics[width=0.98\columnwidth]{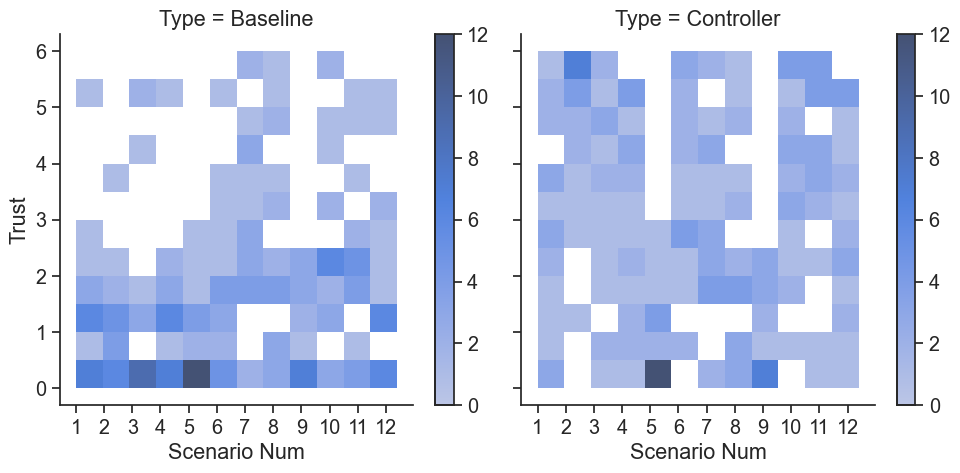}
  \vspace{-3mm}
  \caption{The figures show participants perceived robots using CONFIDANT as more trustworthy (right) compared to baseline robots (left). The heatmap shows the densities of the 7-point likert responses for the 12 test set scenarios.}
  \label{fig:phase2_trust}
\end{figure}

\begin{figure}[!t]
  \centering
  \includegraphics[width=0.98\columnwidth]{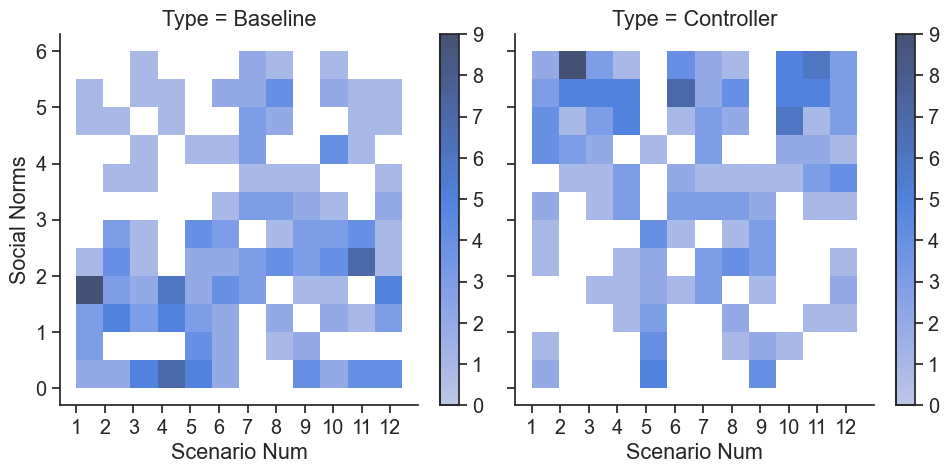}
  \vspace{-3mm}
  \caption{The figures show participants perceived robots using CONFIDANT as more socially-aware (right) compared to baseline robots (left). The heatmap shows the densities of the 7-point likert responses for the 12 test set scenarios.}
  \label{fig:phase2_social_norms}
\end{figure}

To understand whether \controller truly improves users' perceptions, we generated video recordings of conversation scenarios. Participants were asked to respond to a questionnaire that measured the robot's ability to follow appropriate \textit{social norms}, its ability to follow appropriate \textit{privacy norms}, and their level of \textit{trust} toward the robot. 
Our ANCOVA analysis for the social norms scale demonstrated that participants found the robot to more strongly follow expected norms of human conversation when it used our controller ($M=3.769$, $SD=1.707$) over the baseline behavior ($M=2.129$, $SD=1.580$), $F(1:86.92)=250.90$, $p<.0001$. Similarly, analysis of the privacy norms scale showed participants rated the appropriateness of the robot's privacy norms significantly higher when it used our controller ($M=3.663$, $SD=1.849$) over the baseline ($M=1.697$, $SD=1.487$), $F(1:86.54)=371.935$, $p<.0001$. Finally, participants regarded their trust towards the robot higher when the robot used our controller ($M=2.872$, $SD=1.901$) over the baseline ($M=1.505$, $SD=1.509$), $F(1:86.47)=176.907$, $p<.0001$. \cref{fig:phase2_social_norms,fig:phase2_privacy,fig:phase2_trust} contain the responses used in our analysis of the robot's abilities. These differences in users' perceptions are more evident in scenarios with sensitive details compared to scenarios without.


Overall, participants perceived the robot as more socially-aware, privacy-aware, and trustworthy, when our privacy controller governed its disclosures. The controller's presence improved user perceptions of the robot's social awareness by 77\%, privacy awareness by 116\%, and trustworthiness by 91\% over the baseline. After filtering the responses for privacy-conscious participants, the improvement to privacy-awareness provided by \controller grows to 154\%. For non-privacy-conscious participants, this improvement is 79\% (\cref{table:scores}).

\section{Discussion}

Conversational agents that maintain the privacy of their users can serve as more effective and trustworthy conversational partners. Sannon et al.~\cite{sannon2020conversationalagents} explore how people's privacy perceptions of chatbots change when CPM theory guides the chatbot's social interactions and data-sharing practices. However, for social robots, maintaining the privacy of users is more complex due to the dynamic settings in which they function, such as team interactions in the workplace \cite{traeger2020vulnerable}, or overhearing conversations between parents and children \cite{cagiltay2020investigating}.  Our findings provide preliminary evidence for the effectiveness of \controller at managing privacy in social robots. Although our evaluations are performed with social robots, this privacy controller design may extend to other conversational agents. However, privacy typically comes at the expense of another attribute in the form of a trade-off~\cite{wright2002privacy}. While our approach at managing privacy does not negatively impact the perceived social awareness and trustworthiness of the robot, it may come at the expense of usability in certain circumstances. The following sections detail the drawbacks of \controller.

\subsection{Limitations}

While our work shows promising results, it must be considered alongside potential limitations. First, some scenarios created to train our privacy controller are subjective (i.e., participant responses varied significantly depending on participants). In particular, scenarios categorized as ``privacy violations for the greater good'' are determined based on what is considered morally right. While other category types centering around rules and responsibilities can be interpreted based on strict policies or legal boundaries, ethics are influenced by personal philosophies and belief systems. Ethical standards we enforce in the robot may not be universally recognized among all individuals. Second, we excluded highly sensitive conversations (e.g., related to abuse, substance use, or severe medical ailments) from our dataset to minimize psychological risk to our participants. Without these topics, we are unable to assess privacy in such cases. Third, our 58 scenarios only provides partial coverage of the vast space of conversation topics and dynamics. Finally, the current design of our privacy controller only accounts for the context available within a conversation and does not consider additional context such as pre-existing relationships or users' privacy preferences, which may otherwise impact sensitivity determinations.

\subsection{Future Work} \label{subsec:future_work}

The scope of this work is limited to the design and evaluation of a generic ``one size fits all'' privacy controller. To address the system design limitations discussed in the previous section, our future work aims to implement adaptive and personalized features such as speaker relationships and conversation location. Additionally, we plan to explore how other factors highlighted by prior work, including personalization, ownership of devices, social roles, ownership of conversation content, dynamic relationships, co-embodiment, and social context~\cite{luria2020social,reig2021social,reig2020not}, can be integrated into \controller. We also plan to extract more user-specific context when gathering and sharing information~\cite{luria2020social}. \textit{Speaker relationships} will be registered by allowing each user to specify their relationships with other users during the initial robot setup and make any necessary changes moving forward. \textit{Conversation locations} will be implemented using Visual Place Recognition (VPR) to classify images of rooms and identify the location where the conversation takes place (e.g., in a home~\cite{urvsivc2016part} or non-domestic environments such as offices, medical facilities, and restaurants~\cite{lowry2015visual}). Furthermore, incorporating the comprehension and consideration of \textit{non-verbal cues} increases the range of scenarios a social robot can handle and prevents non-verbal data leakage~\cite{mutlu2009nonverbal}. Additional use cases will also be tested to extend \controller to many contexts such as a personal assistant robot in the workplace or a medical robot that interacts with doctors and patients. The robot can adaptively learn new rules specific to these use cases by utilizing user feedback to the robot behavior (e.g., ``\textit{Please don't share this information with anyone again}"). Finally, the issue of scaling the rule generation system can be addressed by expanding our dataset through automated detection of sensitive conversations in online transcripts.

\section{Conclusion}
Successful social integration of robots into daily environments such as in homes or workplaces is important to facilitate human-robot interactions. In these settings, social robots should promote trust and maintain the privacy of those they interact with. In the pursuit of designing more trustworthy robots, we created a privacy controller under simplified assumptions and evaluated its performance in human-robot conversations. Our findings present preliminary evidence for the effectiveness of the privacy controller in creating more privacy-aware, trustworthy, and socially aware social robots.%

\section*{Acknowledgements}
We would like to thank Christine Lee for her help in filming the human-robot interaction videos, the anonymous reviewers for their helpful suggestions, and study participants to their contributions to our data collection.

\newpage

\balance
\printbibliography
\end{document}